\documentclass[runningheads]{llncs}
\usepackage{graphicx}

\usepackage{tikz}
\usepackage{comment} 
\usepackage{amsmath,amssymb} 
\usepackage{color}
\usepackage{multirow}

\begin{document}
\pagestyle{headings}
\mainmatter
\def\ECCVSubNumber{415}  

\title{Identity-Guided Human Semantic Parsing for Person Re-Identification} 

\titlerunning{Identity-Guided Human Semantic Parsing for Person Re-Identification}
%
\author{Kuan Zhu\inst{1,2}\and
Haiyun Guo\inst{1}\and
Zhiwei Liu\inst{1,2}\and
Ming Tang\inst{1,3}\and
Jinqiao Wang\inst{1,2}}
\authorrunning{Zhu et al.}
%
\institute{National Laboratory of Pattern Recognition, Institute of Automation,\\Chinese Academy of Sciences, Beijing, China \and
School of Artificial Intelligence, University of Chinese Academy of Sciences,\\Beijing, China \and
Shenzhen Infinova Limited, Shenzhen, China
 \\
\email{\{kuan.zhu, haiyun.guo, zhiwei.liu, tangm, jqwang\}@nlpr.ia.ac.cn}}
\maketitle


\begin{abstract}
Existing alignment-based methods have to employ the pre-trained human parsing models to achieve the pixel-level alignment, and cannot identify the personal belongings (e.g., backpacks and reticule) which are crucial to person re-ID. In this paper, we propose the identity-guided human semantic parsing approach (ISP) to locate both the human body parts and personal belongings at pixel-level for aligned person re-ID only with person identity labels. We design the cascaded clustering on feature maps to generate the pseudo-labels of human parts. Specifically, for the pixels of all images of a person, we first group them to foreground or background and then group the foreground pixels to human parts. The cluster assignments are subsequently used as pseudo-labels of human parts to supervise the part estimation and ISP iteratively learns the feature maps and groups them. Finally, local features of both human body parts and personal belongings are obtained according to the self-learned part estimation, and only features of visible parts are utilized for the retrieval. Extensive experiments on three widely used datasets validate the superiority of ISP over lots of state-of-the-art methods. Our code is available at \url{{\textcolor{blue}{https://github.com/CASIA-IVA-Lab/ISP-reID}}}.
\end{abstract}

\section{Introduction}
Person re-identification (re-ID), which aims to associate the person images captured by different cameras from various viewpoints, has attracted increasing attention from both the academia and the industry. However, the task of person re-ID is inherently challenging because of the ubiquitous misalignment issue, which is commonly caused by part occlusions, inaccurate person detection, human pose variations or camera viewpoints changing. All these factors can significantly change the visual appearance of a person in images and greatly increase the difficulty of this retrieval  problem.

\begin{figure}
\centering
\includegraphics[width=10cm, height=5.7cm]{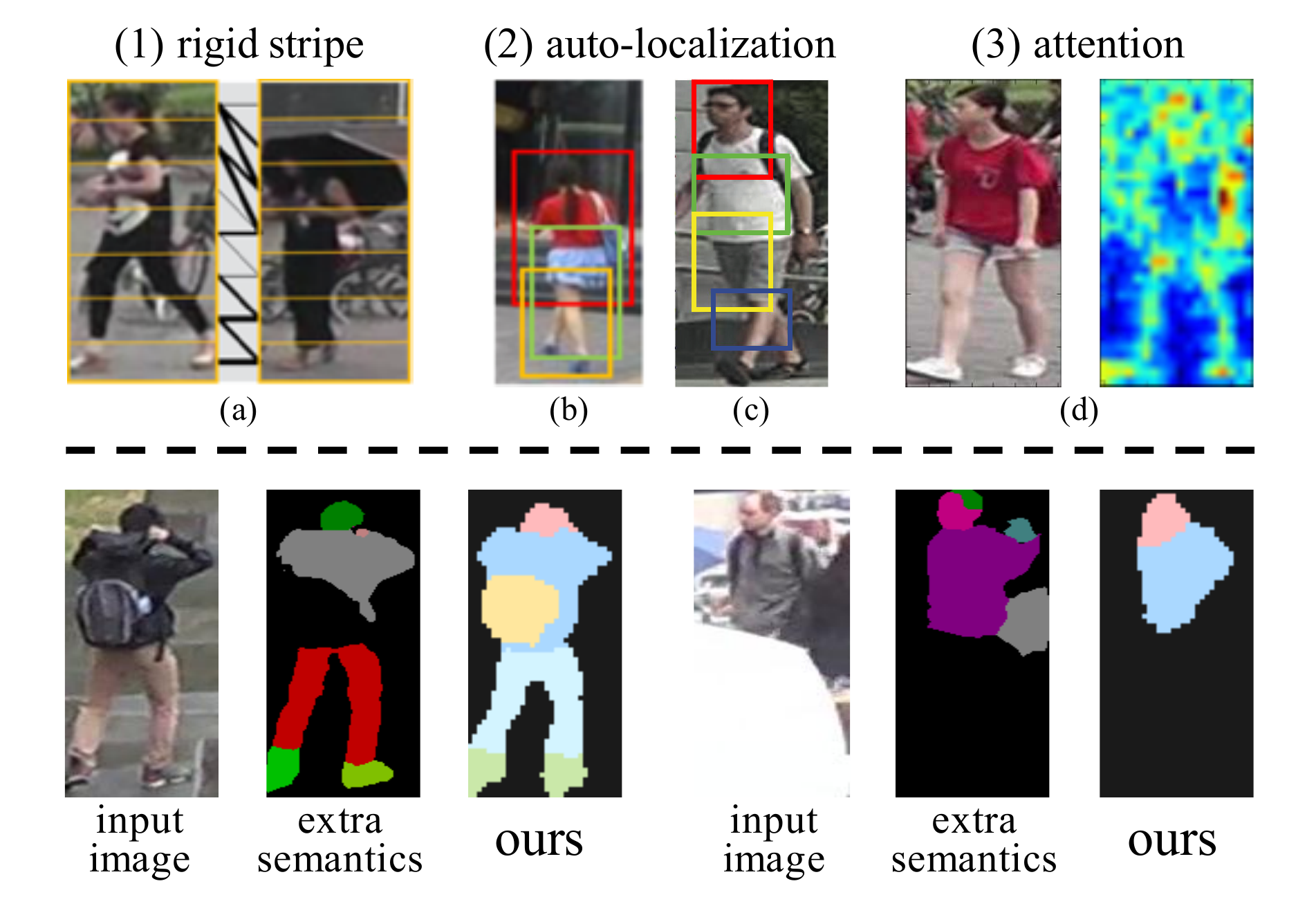}
\caption{The alignment-based methods. From (a) to (d): AlignedReID \cite{AlignedReID}, MSCAN \cite{MSCAN}, DPL \cite{DPL}, MHN \cite{MHN}. The extra semantic in the second row is predicted by the pre-trained parsing model \cite{SCHP}, which exclude the personal belongings and are error-prone when one person is occluded by another. Our method is the first extra semantic free method which can locate both the human body parts and personal belongings at pixel-level, and explicitly identify the visible parts in an occluded image}
\label{fig:introduction}
\end{figure}

In recent years, plenty of efforts have been made to alleviate the misalignment problem. The extra semantic free methods try to address the misalignment issue through a self-learned style. However, they can only achieve coarse alignment at region-level. These methods could be roughly summarized to the following streams:  
(1) The rigid stripe based methods, which directly partition the person image into fixed horizontal stripes \cite{PCB,MGN,AlignedReID,Pyramid}.
(2) The auto-localization based methods, which try to locate human parts through the learned grids \cite{MSCAN,DPL,PAR}. 
(3) The attention based methods, which construct the part alignment through enhancing the discriminative regions and suppressing the background \cite{HA-CNN,MHN,CAMA,CASN}. 
Most of the above methods are coarse with much background noise in their located parts and do not consider the situation that some human parts disappear in an image due to occlusion. The first row of Figure \ref{fig:introduction} illustrates these streams.

The extra semantic based methods inject extra semantic in terms of part/pose to achieve the part alignment at pixel-level \cite{SPReID,LIP,MGCAM,DSA-reID}. Their success heavily counts on the accuracy of the extra pre-trained human parsing models or pose estimators. Most importantly, the identifiable personal belongings (e.g., backpacks and reticule), which are the potentially useful contextual cues for identifying a person, cannot be recognized by these pre-trained models and discarded as background. The failure cases of the extra semantic based methods are shown in the second row of Figure \ref{fig:introduction}.

In this paper, we propose an extra semantic free method, Identity-guided Semantic Parsing (ISP), which can locate both human body parts and potential personal belongings at pixel-level only with the person identity labels. 
Specifically, we design the cascaded clustering on feature maps and regard the cluster assignments as the pseudo-labels of human parts to supervise the part estimation. For the pixels of all images of a person, we first group them to foreground or background according to their activations on feature maps, basing on the reasonable assumption that the classification networks are more responsive to the foreground pixels than the background ones \cite{residual_attn,CBAM,weakly2}. In this stage, the foreground parts are automatically searched by the network itself rather than manually predefined, and the self-learned scheme can capture the potentially useful semantic of both human body parts and personal belongings.  

Next, we need to assign the human part labels to the foreground pixels. The difficulty of this stage lies in how to guarantee the semantic consistency across different images in terms of the appearance/pose variations, and especially the occlusion, which has not been well studied in previous extra semantic free approaches. To overcome this difficulty, we cluster the foreground pixels of all the images with the same ID, rather than those of a single image, into human parts (e.g., head, backpacks, upper-body, legs and shoes), so that the number of assigned semantic parts of a single image can adaptively vary when the instance is occluded. Consequently, our scheme is robust to the occlusion and the assigned pseudo-labels of human parts across different images are ensured to be semantically consistent. The second row of Figure \ref{fig:introduction} shows the assigned pseudo-labels.

We iteratively cluster the pixels of feature maps and employ the cluster assignments as pseudo-labels of human parts to learn the part representations. In this iterative mechanism, the generated pseudo-labels become finer and finer, resulting in the more and more accurate part estimation. The predicted probability maps of part estimation are then used to conduct the part pooling for partial representations of both human body parts and personal belongings. During matching, we only consider local features of the shared-visible parts between probe and gallery images. Besides, ISP is a generally applicable and backbone-agnostic approach, which can be readily applied in popular networks. 

We summarize the contributions of this work as follows:
\begin{itemize}
\item In this paper, we propose the identity-guided human semantic parsing approach (ISP) for aligned person re-ID, which can locate both the human body parts and personal belongings (e.g., backpacks and reticule) at pixel-level only with the image-level supervision of person identities.
\item To the best of our knowledge, ISP is the first extra semantic free method that can explicitly identify the visible parts from the occluded images. The occluded parts are excluded and only features of the shared-visible parts between probe and gallery images are considered during the feature matching.
\item We set the new state-of-the-art performance on three person re-ID datasets, Market-1501 \cite{Market1501}, DukeMTMC-reID \cite{DukeMTMC-reID} and CUHK03-NP \cite{CUHK03-1,CUHK03-2}. 
\end{itemize}


\section{Related work}

\subsection{Semantic learning with image-level supervision}
To the best of our knowledge, there is no previous work to learning human semantic parsing with image-level supervision but only weakly-supervised methods for semantic segmentation \cite{weakly1,weakly2,weakly3,weakly4,weakly5,weakly6,weakly7,weakly8}, which aim to locate objects like person, horse or dog at pixel-level with image-level supervision. However, all these methods cannot be used for the weakly-supervised human parsing task because they focus on different levels. Besides, their complex network structures and objective functions are not suitable for the end-to-end learning of person re-ID. Therefore, we draw little inspiration from these methods.

\subsection{Alignment-based person re-ID}
The alignment-baed methods can be roughly summarized to the four streams:

\textbf{Rigid stripe based approaches.} Some researchers directly partition the person image into rigid horizontal stripes to learn local features \cite{PCB,MGN,AlignedReID,Pyramid}. Wang et al. \cite{MGN} design a multiple granularity network, which contains horizontal stripes of different granularities. Zhang et al. \cite{AlignedReID} introduce a shortest path loss to align rigidly divided local stripes. However, the stripe-based partition is too coarse to well align the human parts and introduces lots of background noise.

\textbf{Auto-localization based approaches.} A few works have been proposed to automatically locate the discriminative parts by incorporating a regional selection sub-network \cite{MSCAN,DPL}. Li et al. \cite{MSCAN} exploit the STN \cite{STN} for locating latent parts and subsequently extract aligned part features. 
However, the located grids of latent parts are still coarse and with much overlap. Besides, they produce a fixed number of latent parts, which cannot handle the occluded images.

\textbf{Attention based approaches.} Attention mechanism constructs alignment by suppressing background noise and enhancing the discriminative regions \cite{MHN,HA-CNN,DuATM,Mancs,CAMA,CASN}. However, these methods cannot explicitly locate the semantic parts and the consistency of focus area between images is not guaranteed. 

\textbf{Extra semantic based approaches.} Many works employ extra semantic in terms of part/pose to locate body parts \cite{P2-Net,PoseTransfer,PGFA,PSE,MGCAM,PDC,AANet,GLAD,AACN,PoseInvariant} and try to achieve the pixel-level alignment. Kalayeh et al. \cite{SPReID} propose to employ pre-trained human parsing model to provide extra semantic. Zhang et al. \cite{DSA-reID} further adopt DensePose \cite{DensePose} to get densely semantic of 24 regions for a person. However, the requiring of extra semantic limits the utility and robustness of these methods. First, the off-the-shelf models can make mistakes in semantic estimation and these methods cannot recorrect the mistakes throughout the training. Second, the identifiable personal belongings like backpacks and reticule, which are crucial for person re-ID, cannot be recognized and ignored as background. 



In this paper, we adopt the clustering to learn the human semantic parsing only with person identity labels, which can locate both human body parts and personal belongings at pixel-level. Clustering is a classical unsupervised learning method that groups similar features, while its capability has not been fully explored in the end-to-end training of deep neural networks. Recently, Mathilde et al. \cite{DeepCluster} adopt clustering to the end-to-end unsupervised learning of image classification. 
Lin et al. \cite{BUC}  also use clustering to solve the unsupervised person re-ID task. Different from them, we go further by grouping pixels to human parts to generate the pseudo-part-labels at pixel-level, which is more challenging due to various noises. Moreover, the results of clustering must guarantee the semantic consistency across images. 

\section{Methodology}

\begin{figure*}[t]
\centering
\includegraphics[width=1.0\linewidth, height=6cm]{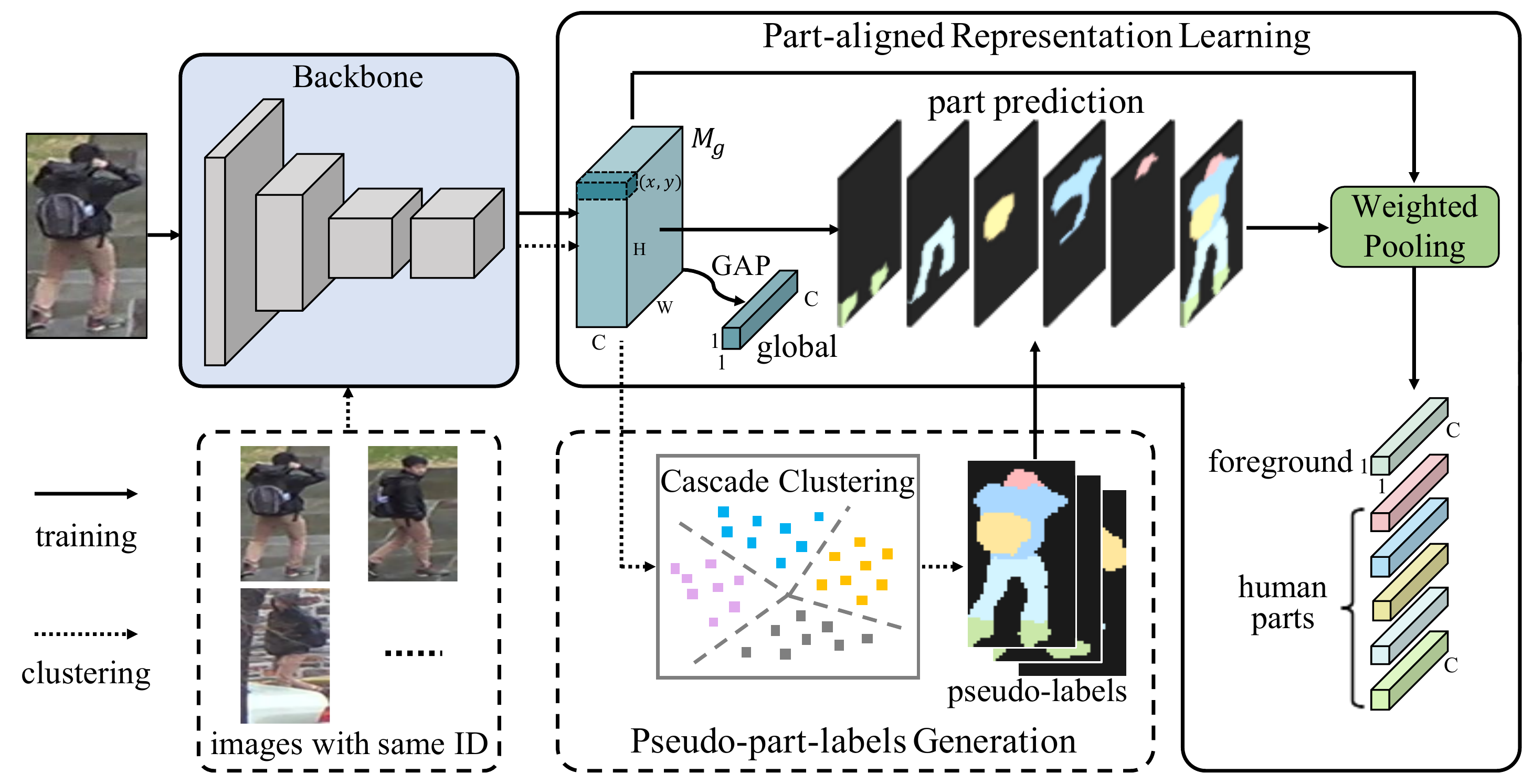}
\caption{The overview of ISP. The {\it solid line} represents the training phase and the {\it dotted line} represents the clustering phase. The two stages are iteratively done until the network converges. ISP is a generally applicable and backbone-agnostic approach.}
\label{fig:overview}
\end{figure*}

The overview of ISP is shown in Figure \ref{fig:overview}. There are mainly two processes in our approach, i.e., pseudo-part-labels generation and part-aligned representation learning. We repeat the above two processes until the network converges.

\subsection{Pixel-level part-aligned representation learning}
Given $n$ training person images $\{X_i\}_{i=1}^n$ from $n_{id}$ distinct people and their identity labels $\{y_i\}^n_{i=1}$ (where $y_i \in \{1,...,n_{id}\}$), we could learn the human semantic parsing to obtain the pixel-level part-aligned representations for person re-ID. For image $x_i$, the backbone mapping function (defined as $f_{\theta}$) will output the global feature map:
\begin{equation}
M_g^{c\times h\times w}=f_{\theta}(x_i)
\end{equation}
where $\theta$ is the parameters of backbone, and $c, h, w$ is the channel, height and width. For clear exposition, we omit the channel dimension and denote by $M_g(x,y)$ the feature at spatial position $(x,y)$, which is a vector of $c$-dim.

The main idea of our pixel-level part-aligned representations is to represent human parts with the representations of pixels belonging to that part, which is the aggregation of the pixel-wise representations weighted by a set of confidence maps. Each confidence map is used to surrogate a human part.
Assuming there are $K-1$ human parts and one background part in total, we need to estimate $K$ confidence maps of different semantic parts for every person image. It should be noted that we treat the personal belongings as one category of human parts. The $K$ confidence maps is defined as $P_0, P_1,..., P_{K-1}$, where each confidence map $P_k$ is associated with a semantic part. We denote by $P_k(x,y)$ the confidence of pixel $(x,y)$ belonging to semantic part $k$.  Then the feature map of part $k$ can be extracted from the global feature map by:
\begin{equation}
M_k=P_k\circ M_g
\end{equation}
where $k\in\{0,...,K-1\}$ and $\circ$ is the element-wise product. Adding $M_k$ from $k=1$ to $k=K-1$ in element-wise will get the foreground feature map $M_{f}$.  Ideally, for the occluded part $k$ in an occluded person image, $\forall_{(x,y)} P_k(x,y)=0$ should be satisfied, which is reasonable that the network should not produce representations for the invisible parts.

\subsection{Cascaded clustering for pseudo-part-labels generation}

Existing studies integrate human parsing results to help capture the human body parts at pixel-level \cite{SPReID,LIP,MGCAM}. However, there are still many useful contextual cues like backpacks and reticule that do not fall into the scope of manually predefined human body parts. 
We design the cascaded clustering on feature maps $M_g$ to generate the pseudo-labels of human parts, which includes both human body parts and personal belongings.

Specifically, in the first stage, for all $M_g$ of the same person, we group their pixels into the foreground or background according to the activation, basing on the conception that the foreground pixels have a higher response than background ones \cite{residual_attn,CBAM,weakly2}. In this stage, the discriminative foreground parts are automatically searched by the network and the self-learned scheme could apply both the human body parts and the potential useful personal belongings with high response. 
We regard the $l_2$-norm of $M_g(x,y)$ as the activation of pixel $(x,y)$. For all pixels of a $M_g$, we normalize their activations with their maximum: 
\begin{equation}
a(x,y)=\frac{||M_g(x,y)||_2}{\max_{(i,j)}||M_g(i,j)||_2}	
\end{equation}
where $(i,j)$ is the positions in the $M_g$ and the maximum of $a(x,y)$ equals to 1.

In the second stage, we cluster all the foreground pixels assigned by the first clustering step into $K\!-\!1$ semantic parts. The number of semantic parts for a single image could be less than $K\!-\!1$ when the person is occluded because the cluster samples are foreground pixels of all $M_g$ from the images of the same person, rather than $M_g$ of a single image. Therefore, the clustering is robust to the occlusion  and the part assignments across different images are ensured to be semantically consistent.
In this stage, we focus on the similarities and differences between pixels rather than activation thus $l_2$-normalization is used: 
\begin{equation}
D(x,y)=\frac{M_g(x,y)}{||M_g(x,y)||_2}
\end{equation}

The cluster assignments are then used as the pseudo-labels of human parts, which contain the personal belongings as one foreground part, to supervise the learning of human semantic parsing. We assign label 0 to background and the body parts are assigned to label $\{1,...,K-1\}$ according to the average position from top to down. ISP iteratively does the cascade clustering on feature maps and uses the assignments as pseudo-part-labels to learn the partial representations. In this iterative mechanism, the generated pseudo-labels become finer and finer, resulting in more and more accurate part estimation for aligned person re-ID.

\textbf{Optimization.} For part prediction, we use a linear layer followed by softmax activation as the classifier, which is formulated as:

\begin{equation}
P_k(x,y)=softmax(W_k^TM_g(x,y))=\frac{{\rm exp}(W_k^TM_g(x,y))}{\sum_{i=0}^{K-1}{\rm exp}(W_i^TM_g(x,y))}
\end{equation}
where $k\in\{0,...,K-1\}$ and $W$ is the parameters of linear layer. 

We assign the probability $P_k(x,y)$ as the confidence of pixel $(x,y)$ belonging to the semantic part $k$ and employ cross-entropy loss to optimize the classifier:
\begin{equation}
\mathcal{L}_{parsing}=\sum_{x,y}-\log P_{k_i}(x,y)
\end{equation}
where $k_i$ is the generated pseudo-label of human parts for pixel $(x,y)$.

\subsection{Objective function}
The representation for semantic part $k$ is obtained by $F_k=GAP(M_k)$, where $GAP$ means global average pooling. 
We concatenate all $F_k$ except $k\!=\!0$ and regard the outcome as a whole representation of local parts for training. Besides, the representations for foreground and global image are directly obtained by $F_f\!=\!GAP(M_f)$, $F_g\!=\!GAP(M_g)$. In fact, the probability map product together with the GAP is the operation of weighted pooling as indicated in Figure \ref{fig:overview}. 

In the training phase, we employ three groups of basic losses for the representations of local part, foreground and global image separately, which are denoted as $\mathcal{L}_{p}$, $\mathcal{L}_{f}$ and $\mathcal{L}_{g}$. For each basic loss group, we follow \cite{baseline} to combine the triplet loss \cite{triplet_loss} and cross-entropy loss with label smoothing \cite{label_smooth}. Therefore, the overall objective function is: 
\begin{equation}
\mathcal{L}_{reid}=\mathcal{L}_{p}+\mathcal{L}_{f}+\mathcal{L}_{g}+\alpha \mathcal{L}_{parsing}
\end{equation}
where $\alpha$ is the balanced weight and is set to $0.1$ in our experiments.

\begin{figure}
\centering
\includegraphics[height=4cm]{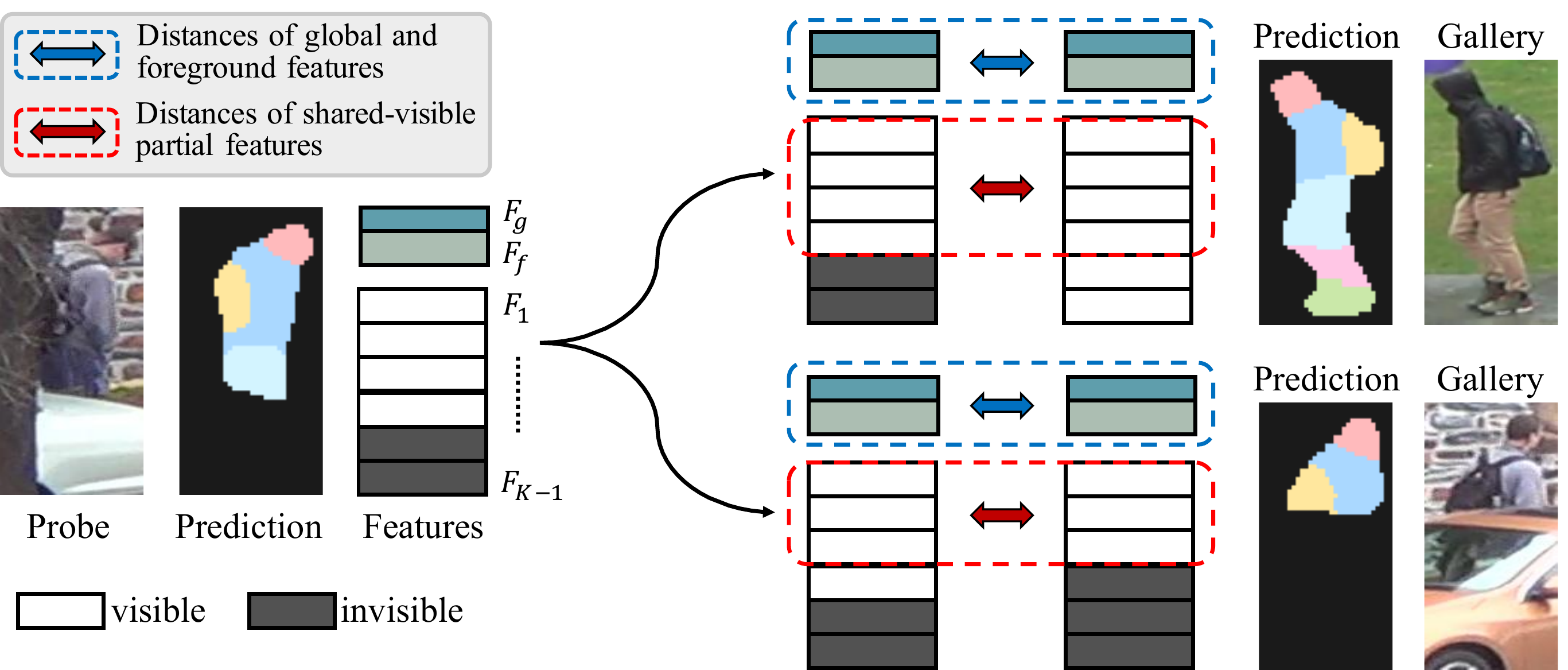}
\caption{The matching strategy of ISP. The distance between probe and gallery images are measured by features of the global image and foreground part, which always exist, and the features of shared-visible parts.}
\label{fig:matching}
\end{figure}

\subsection{Aligned representation matching}

As illustrated in Figure \ref{fig:matching}, the final distance between query and gallery images consists of two parts. One is the distance of global and foreground features, which always exist. The other is the distance of the partial features between the shared-visible human parts. The matching strategy is inspired by \cite{PGFA}, but \cite{PGFA} utilizes extra pose information and only achieves stripe-level alignment, while we do not require any extra semantic and could identify the visible parts at pixel-level. As the ${\rm argmax_i}P_i(x,y)$ indicates the semantic part of pixel $(x,y)$ belonging to, we could easily obtain the label of whether part $k$ is visible $l_k\in\{0,1\}$ by:
\begin{equation}
l_k=
\left\{
\begin{aligned}
&1, ~~~~{\rm if}~\exists (x,y)\!\in\!\{(x,y)|{\rm argmax_i}{P_i(x,y)}=k\}\\
&0, ~~~~{\rm else}\\
 \end{aligned}
 ~(i=0,...,K-1)
 \right.
\end{equation}
Now the distance $d_k$ of the $k$th part between query and gallery images is:\begin{equation}
	d_k=D(F_k^q,F_k^g)~~(k=1,...,K-1)
\end{equation}
where $D()$ denotes the distance metric, which is cosine distance in this paper. $F_k^q,F_k^g$ denote the $k$th partial feature of the query and gallery image, respectively. Similarly, the measure distance between global and foreground features are formulated as: $d_{g}=D(F_g^q,F_g^g)$, $d_f=D(F_f^q,F_f^g)$. Then, the final distance $d$ could be obtained by:
\begin{equation}
	d=\frac{\sum^{K-1}_{k=1}(l_k^q\cdot l_k^g)d_k+(d_g+d_f)}{\sum^{K-1}_{k=1}(l_k^q\cdot l_k^g)+2}
\end{equation}
If the $k$th parts of both the query and gallery images are visible, $l_k^q\cdot l_k^g=1$. Else, $l_k^q\cdot l_k^g=0$. To the best of our knowledge, ISP is the first extra semantic free method that explicitly addresses the occlusion problem.

\section{Experiments}
\subsection{Datasets and evaluation metrics}

\textbf{Holistic person re-ID datasets.} We select three widely used holistic person re-ID benchmarks, Market-1501 \cite{Market1501} which contains 32668 person images of 1501 identities, DukeMTMC-reID \cite{DukeMTMC-reID} which contains 36411 person images of 1402 identities and CUHK03-NP (New Protocol) \cite{CUHK03-1,CUHK03-2} which contains 14097 person images of 1467 identities for evaluation. Following common practices, we use the cumulative matching characteristics (CMC) at Rank-1, Rank-5, and the mean average precision (mAP) to evaluate the performance. 

\noindent\textbf{Occluded person re-ID datasets.} We also evaluate the performance of ISP in the occlusion scenario. Occluded-DukeMTMC \cite{PGFA}, which contains 15618 training images, 17661 gallery images, and 2210 occluded query images, is by far the largest and the only occluded person re-ID datasets that contains training set. It is a new split of DukeMTMC-reID \cite{DukeMTMC-reID} and the training/query/gallery set contains $9\%/100\%/10\%$ occluded images, respectively. 
Therefore, we demonstrate the effectiveness of ISP in occluded scenario on this dataset. 

\subsection{Implementation details}


\textbf{Data preprocessing.} The input images are resized to 256$\times$128 and the global feature map $M_g$ is $1/4$ of the input size. As for data augmentation, we adopt the commonly used random cropping \cite{random_earse2}, horizontal flipping and random erasing \cite{Mancs,random_earse2,random_earse3} (with a probability of 0.5) in both the baseline and our schemes.
 
\noindent\textbf{Optimization.} 
The backbone network is initialized with the pre-trained parameters on ImageNet \cite{imagenet}. We warm up the model for 10 epochs with a linearly growing learning rate from $3.5\times10^{-5}$ to $3.5\times10^{-4}$. Then, the learning rate is decreased by a factor of $0.1$ at $40$th and $70$th epoch. We observe that 120 epochs are enough for model converging. The batch size is set to 64 and adam method is adopted to optimize the model. All our methods are implemented on PyTorch. 

\noindent\textbf{Clustering for reassignment.} We adopt k-means as our clustering algorithm and reassign the clusters every $n$ epochs, which is a tradeoff between the parameter updating and the pseudo-label generation. We find out that simply setting $n\!=\!1$ is nearly optimal. We do not define any initial pseudo-labels for person images and the first clustering is directly conducted on the feature maps output by the initialized backbone. As for the time consumption, to train a model, the overall clustering time is about 6.3h/5.4h/3.1h for datasets of DukeMTMC-reID/Market1501/CUHK03-NP through multi-processes with one NVIDIA TITAN X GPU. Most importantly, the testing time is not increased at all.

\subsection{Comparison with state-of-the-art methods}

\begin{table}[t]
    \caption{Comparison with state-of-the-art methods of the holistic re-ID problem. The $1^{st}$/$2^{nd}$ results are shown in red/blue, respectively. The methods in the 1st group are rigid stripe based. The methods in the 2nd group are auto-localization based. The 3rd group is attention based methods. The methods in the 4th group are extra semantic based. The last line is our method
    }
    \centering
    \small
    \begin{tabular}{lcccccccccccccc}
      \hline 
      \multirow{3}*{Methods}&\multirow{3}*{Ref}&~&\multicolumn{3}{c}{\multirow{2}*{DukeMTMC}}&~&\multicolumn{3}{c}{\multirow{2}*{Market1501}}&~&\multicolumn{4}{c}{CUHK03-NP} \\ \cline{12-15}
      ~&~&~&~&~&~&~&~&~&~&~&\multicolumn{2}{c}{Labeled}&\multicolumn{2}{c}{Detected}\\ 
      \cline{3-15}
      ~&~&~&R-1&R-5&mAP&~&R-1&R-5&mAP&~&R-1&mAP&R-1&mAP\\ 
      
      \hline
      AlignedReID \cite{AlignedReID}&Arxiv18&~&-&-&-&~&91.8&97.1&79.3&~&-&-&-&- \\
      PCB+RPP \cite{PCB}&ECCV18&~&83.3&-&69.2&~&93.8&97.5&81.6&~&-&-&63.7&57.5\\
      MGN \cite{MGN}&MM18&~&88.7&-&\textcolor{blue}{78.4}&~&\textcolor{red}{95.7}&-&86.9&~&68.0&67.4&66.8&66.0\\ 
      \hline
      MSCAN \cite{MSCAN}&CVPR17&~&-&-&-&~&80.8&-&57.5&~&-&-&-&-\\
      PAR \cite{PAR}&ICCV17&~&-&-&-&~&81.0&92.0&63.4&~&-&-&-&- \\
      \hline 
      DuATM \cite{DuATM}&CVPR18&~&81.8&90.2&64.6&~&91.4&97.1&76.6&~&-&-&-&-\\
      Mancs \cite{Mancs}&ECCV18&~&84.9&-&71.8&~&93.1&-&82.3&~&69.0&63.9&65.5&60.5\\
      IANet \cite{IANet}&CVPR19&~&87.1&-&73.4&~&94.4&-&83.1&~&-&-&-&-\\
      CASN+PCB \cite{CASN}&CVPR19&~&87.7&-&73.7&~&94.4&-&82.8&~&73.7&68.0&71.5&64.4\\ 
      CAMA \cite{CAMA}&CVPR19&~&85.8&-&72.9&~&94.7&98.1&84.5&~&70.1&66.5&66.6&64.2\\ 
      MHN-6 \cite{MHN}&ICCV19&~&\textcolor{blue}{89.1}&\textcolor{blue}{94.6}&77.2&~&95.1&98.1&85.0&~&77.2&72.4&71.7&65.4\\
      \hline
      SPReID \cite{SPReID}&CVPR18&~&84.4&-&71.0&~&92.5&-&81.3&~&-&-&-&-\\           
      PABR \cite{PABR}&ECCV18&~&84.4&92.2&69.3&~&91.7&96.9&79.6&~&-&-&-&-\\       
      AANet \cite{AANet}&CVPR19&~&87.7&-&74.3&~&93.9&-&83.4&~&-&-&-&-\\
      DSA-reID \cite{DSA-reID}&CVPR19&~&86.2&-&74.3&~&\textcolor{red}{95.7}&-&\textcolor{blue}{87.6}&~&\textcolor{red}{78.9}&\textcolor{red}{75.2}&\textcolor{red}{78.2}&\textcolor{red}{73.1}\\
      $P^2$-Net \cite{P2-Net}&ICCV19&~&86.5&93.1&73.1&~&95.2&\textcolor{blue}{98.2}&85.6&~&\textcolor{blue}{78.3}&73.6&74.9&68.9\\
      PGFA \cite{PGFA}&ICCV19&~&82.6&-&65.5&~&91.2&-&76.8&~&-&-&-&-\\
      \hline
      ISP (ours)&ECCV20&~&\textcolor{red}{89.6}&\textcolor{red}{95.5}&\textcolor{red}{80.0}&~&\textcolor{blue}{95.3}&\textcolor{red}{98.6}&\textcolor{red}{88.6}&~&76.5&\textcolor{blue}{74.1}&\textcolor{blue}{75.2}&\textcolor{blue}{71.4} \\ 
      \hline
    \end{tabular}
        \label{table:state_of_the_art}
\end{table}

We compare our method with the state-of-the-art methods for holistic and occluded person re-ID in Table \ref{table:state_of_the_art} and Table \ref{table:state_of_the_art_occluded}, respectively.

\textbf{DukeMTMC-reID.} ISP achieves the best performance and outperforms others by at least $0.5\%/1.6\%$ in Rank-1/mAP. On this dataset, the semantic extracted by pre-trained model is error-prone \cite{DSA-reID}, which leads to significant performance degradation for extra semantic based methods. This also proves the learned semantic parts are superior to the outside ones in robustness.

\textbf{Market1501.} ISP achieves the best performance on mAP accuracy and the second best on Rank-1. We further find that the improvement of mAP score brought by ISP is larger than that of Rank-1, which indicates ISP effectively advances the ranking positions of misaligned person images as mAP is a comprehensive index that considers all the ranking positions of the target images.

\textbf{CUHK03-NP.} ISP achieves the second best results. In CUHK03-NP, a great many of images  contain incomplete person bodies and some human parts even disappear from all the images of a person. But ISP requires at least every semantic part appears once for a person to  guarantee a high consistency on semantic. Even so, ISP still outperforms all other approaches except DSA-reID \cite{DSA-reID} which employs extra supervision by a pre-trained DensePose model \cite{DensePose}, while our method learns the pixel-level semantic without any extra supervision.


\setlength{\tabcolsep}{4pt}
\begin{table}
\begin{center}
\caption{Comparison with state-of-the-art methods of the occluded re-ID problem on Occluded-DukeMTMC. Methods in the 1st group are for the holistic re-ID problem. Methods in the 2nd group utilize extra pose information for occluded re-ID problem. The methods in the 3rd group do not adopt extra semantic. The last line is our method}
\label{table:state_of_the_art_occluded}
\small
\begin{tabular}{lcccc}
\hline\noalign{\smallskip}
Methods & Rank-1 & Rank-5 & Rank-10 & mAP\\
\noalign{\smallskip}
\hline
HACNN \cite{HACNN} & 34.4 & 51.9 & 59.4 & 26.0 \\
Adver Occluded \cite{adver_occluded} & 44.5 & - & - & 32.2 \\
PCB \cite{PCB} & 42.6 & 57.1 & 62.9 & 33.7 \\
\hline
Part Bilinear \cite{part_bilinear} & 36.9 & - & - & - \\
FD-GAN \cite{FD-GAN} & 40.8 & - & - & - \\
PGFA \cite{PGFA} & 51.4 & 68.6 & 74.9 & 37.3 \\
\hline
DSR \cite{DSR} & 40.8 & 58.2 & 65.2 & 30.4 \\
SFR \cite{SFR} & 42.3 & 60.3 & 67.3 & 32.0 \\
\hline
ISP$_{w\!/\!o~arm}$ & 59.5 & 73.5 & 78.0 & 51.4 \\
ISP (ours) & \textbf{62.8} & \textbf{78.1} & \textbf{82.9} & \textbf{52.3} \\ 
\hline
\end{tabular}
\end{center}
\end{table}
\setlength{\tabcolsep}{1.4pt}

\textbf{Occluded-DukeMTMC.} ISP sets a new state-of-the-art performance and outperforms others by a large margin, at least $11.2\%/14.3\%$ in Rank-1/mAP. ISP could explicitly identify the visible parts at pixel-level from the occluded images and only the shared-visible parts between query and gallery images are considered during the feature matching, which greatly improves the performance. As shown in Table \ref{table:state_of_the_art_occluded}, the aligned representation matching strategy brings considerable improvement, e.g., 3.3\% of Rank-1 and 0.9\% of mAP.

\subsection{The performance of the learned human semantic parsing}
As there are no manual part labels for the person re-ID datasets, we adopt the state-of-the-art parsing model SCHP \cite{SCHP} pre-trained on Look into Person (LIP) \cite{LIP} to create the ``ground-truth'' of four parts.\footnote{The parts of hat, hair, sunglass, face in LIP is aggregated as the ``ground-truth'' for Head; the parts of left-leg, right-leg, socks, pants are aggregated as Legs; the parts of left-shoe and right-shoe are aggregated as Shoes.} Then we adopt segmentation Intersection over Union (IoU) to evaluate both the accuracy of the pseudo-part-labels of the training set and that of the semantic estimation on the testing set, which are detailed in Table \ref{table:IoU}, and the results are high enough for the IoU evaluation metric. 
We also find an interesting phenomenon that the accuracy of the predicted semantic estimation on testing set is mostly higher than that of the pseudo-part-labels for the training set, which indicates that, with the person re-ID supervision, our part estimator is robust to the false pseudo-labels and obtains an enhanced generalization capability. 
\newcommand{\tabincell}[2]{\begin{tabular}{@{}#1@{}}#2\end{tabular}}
\setlength{\tabcolsep}{4pt}
\begin{table} 
\small
\begin{center}
\caption{The human semantic parsing performance (\%) of ISP $(K\!=\!6)$}
\label{table:IoU}
\begin{tabular}{cccccc}
\hline\noalign{\smallskip}
IoU&Datasets & Foreground & Head & Legs & Shoes \\
\noalign{\smallskip}
\hline
\multirow{3}*{\tabincell{c}{Pseudo-labels\\Accuracy}}&DukeMTMC  & 65.66 & 68.17 & 61.83 & 58.89 \\ 
~&Market1501 & 65.45 & 54.74 & 67.02 & 55.25 \\
~&CUHK03-NP-Labeled & 51.26 & 68.21 & 52.24 & 57.60 \\
\hline
\multirow{3}*{\tabincell{c}{Prediction\\Accuracy}}&DukeMTMC  & 66.94 & 71.35 & 68.02 & 62.60 \\ 
~&Market1501 & 63.44 & 55.78 & 69.10 & 56.32 \\
~&CUHK03-NP-Labeled & 53.51 & 59.96 & 50.14 & 59.08 \\
\hline
\end{tabular}
\end{center}
\end{table}
\setlength{\tabcolsep}{1.4pt}

\begin{figure}
\centering
\includegraphics[height=3.7cm, width=1.0\linewidth]{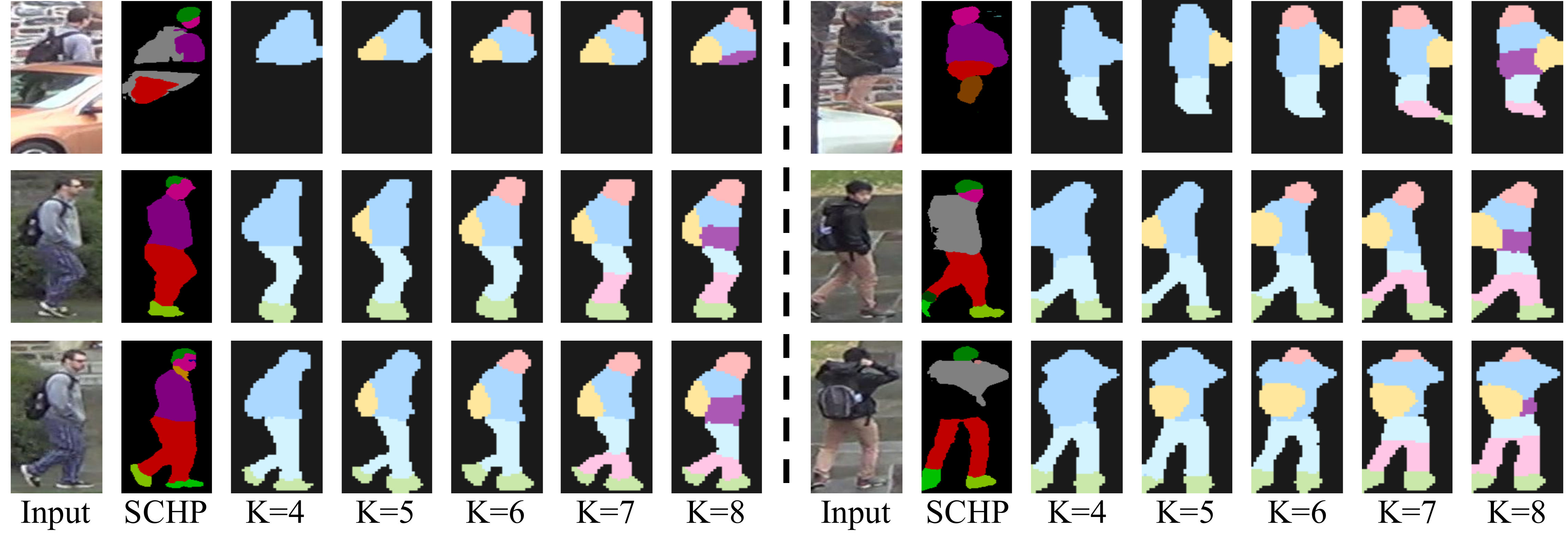}
\caption{The assigned pseudo-labels of human parts. From left to right: input images, semantic estimation by SCHP \cite{SCHP}, the assigned pseudo-labels with different $K$. 
}
\label{fig:all}
\end{figure}

We further conduct three visualization experiments to show the effect of ISP. First, we visualize our pseudo-part-labels under different $K$ and show the comparison with SCHP \cite{SCHP} in Figure \ref{fig:all}, which validates that ISP can recognize the personal belongings (e.g. backpacks and reticule) as one human part while the pre-trained parsing model cannot. 
The first row of Figure \ref{fig:all} shows our capability of explicitly locating the visible parts in occluded images.
Moreover, Figure \ref{fig:all} also validates the semantic consistency in ISP. 

\begin{figure}
\centering
\includegraphics[height=2.05cm, width=1.0\linewidth]{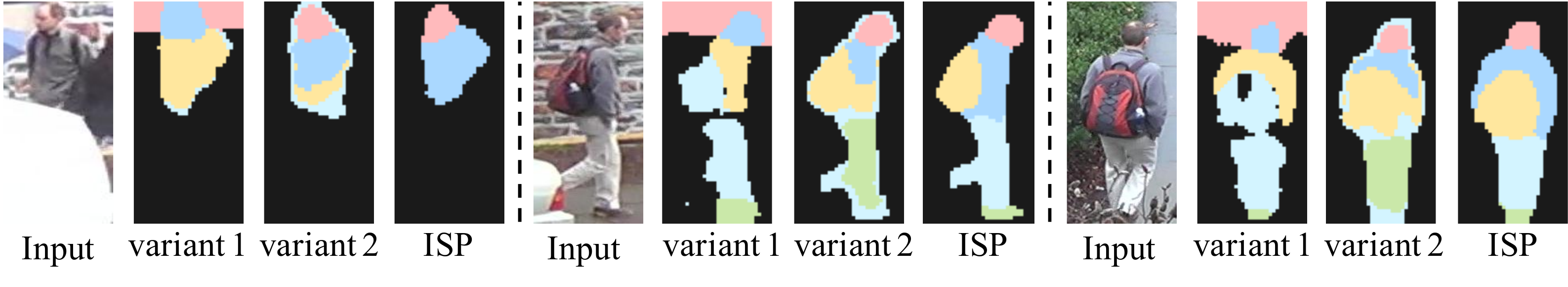}
\caption{The necessity of the operations of cascaded clustering $(K=6)$. 
}
\label{fig:why_cascade}
\end{figure}

Second, to validate the necessity of the cascaded clustering, we compare the pseudo-part-labels by cascaded clustering with Variant 1 which directly clusters the semantic in one step, and Variant 2 which removes the $l_2$ normalization. Figure \ref{fig:why_cascade} indicates that the alignment of Variant 1 is coarse and error-prone, and Variant 2 assigns an unreasonable semantic part with sub-response surround human bodies, which indicates the clustering is influenced by the activation.
 
\begin{figure}
\centering
\includegraphics[height=3cm, width=1.0\linewidth]{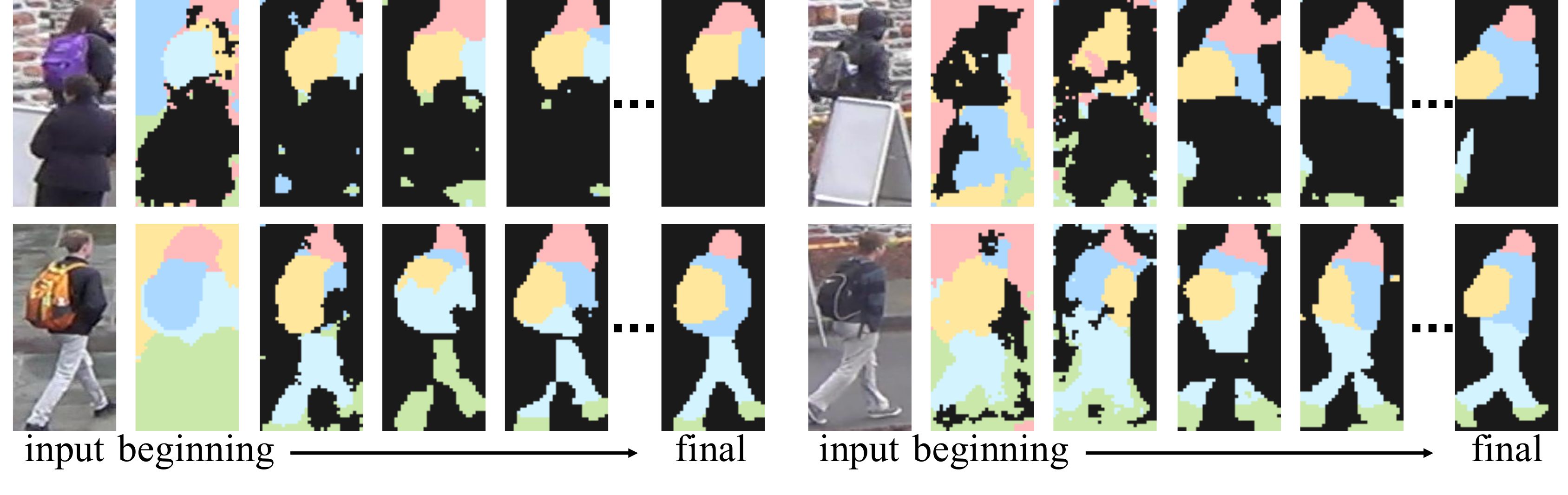} 
\caption{The evolution process of pseudo-labels, which is a gradual process of refinement.}
\label{fig:evolution}
\end{figure}

 Finally, Figure \ref{fig:evolution} shows the evolution process of the pseudo-part-labels $(K\!=\!6)$, which presents a clear process of coarse-to-fine. The first clustering is directly conducted on feature maps output by the initialized network.
 
\subsection{Ablation studies on re-ID performance}
\setlength{\tabcolsep}{4pt}
\begin{table}[h]
\small
\begin{center}
\caption{The ablation studies of $K$. The results show ISP is robust to different $K$}
\label{table:K} \
\begin{tabular}{lccccccccccccc}
      \hline
      \multirow{3}*{K}&~&\multicolumn{3}{c}{\multirow{2}*{DukeMTMC}}&~&\multicolumn{3}{c}{\multirow{2}*{Market1501}}&~&\multicolumn{4}{c}{CUHK03-NP} \\ \cline{11-14}
      ~&~&~&~&~&~&~&~&~&~&\multicolumn{2}{c}{Labeled}&\multicolumn{2}{c}{Detected}\\ \cline{3-14}
    
      ~&~&R-1&R-5&mAP&~&R-1&R-5&mAP&~&R-1&mAP&R-1&mAP\\ 
      \hline
      4 &~&89.6&95.2&79.2&~&\textbf{95.3}&\textbf{98.6}&\textbf{88.6}&~&75.9&72.9&73.5&71.3\\ 
      5 &~&88.6&95.0&79.1&~&94.4&98.2&87.7&~&73.9&72.4&73.1&71.0\\   
      6 &~&89.0&95.1&78.9&~&95.2&98.4&88.4&~&75.9&73.8&\textbf{75.2}&\textbf{71.4}\\   
      7 &~&\textbf{89.6}&\textbf{95.5}&\textbf{80.0}&~&95.0&98.2&88.4&~&\textbf{76.5}&\textbf{74.1}&73.6&70.8\\   
      8 &~&88.9&94.7&78.4&~&94.9&98.5&88.6&~&75.9&73.1&74.0&71.4\\      
\hline
\end{tabular}
\end{center}
\end{table}
\setlength{\tabcolsep}{1.4pt}

\textbf{Choice of $K$ clustering categories.}
Intuitively, the number of clustering centers $K$ determines the granularity of the aligned parts. 
We perform the quantitative ablation studies to clearly find the most suitable $K$. As detailed in Table \ref{table:K}, the performance of ISP is robust to different $K$. Besides, we also find that $\!K=\!5$ is always the worst, which is consistent with its lowest accuracy of semantic parsing. For example, $K\!=\!5$ only obtains the pseudo-labels accuracy (IoU) of 64.25\%, 53.23\% and 54.83\% for foreground, legs and shoes on DukeMTMC-reID. 

\textbf{Learned semantic vs. extra semantic.}
We further conduct experiments to validate the superiority of the learned semantic over extra semantic.
HRNet-W32 \cite{HRNet1} is set as our baseline model. ``+extra info'' means adopting the extra semantic information extracted by SCHP \cite{SCHP} as the human part labels while 
``ISP'' adopts the learned semantic. The results are list in Table \ref{table:ablation_studies}, which show ISP consistently outperforms ``+extra info'' method by a considerable margin. We think this is mainly because: (1) ISP can recognize the identifiable personal belongs while ``+extra info'' cannot. (2) ``+extra info'' cannot recorrect the semantic estimation errors throughout the training while ISP can recorrect its mistakes every epoch, thus ISP is less likely to miss the key clues. 

\setlength{\tabcolsep}{4pt}
\begin{table}
\small
\begin{center}
\caption{The comparison of learned semantic and extra semantic}
\label{table:ablation_studies}
\begin{tabular}{cccccccccccccc}
      \hline
      \multirow{3}*{Model}&~&\multicolumn{3}{c}{\multirow{2}*{DukeMTMC}}&~&\multicolumn{3}{c}{\multirow{2}*{Market1501}}&~&\multicolumn{4}{c}{CUHK03-NP} \\ \cline{11-14}
      ~&~&~&~&~&~&~&~&~&~&\multicolumn{2}{c}{Labeled}&\multicolumn{2}{c}{Detected}\\ \cline{3-14}
    
      ~&~&R-1&R-5&mAP&~&R-1&R-5&mAP&~&R-1&mAP&R-1&mAP\\ 
      \hline
      baseline &~&87.7&94.3&77.2&~&94.0&97.9&85.9&~&71.9&68.5&67.6&64.7\\ 
      +extra info &~&88.6&94.7&79.1&~&94.8&98.4&87.7&~&73.3&71.9&72.2&69.6\\   
      ISP &~&89.6&95.5&80.0&~&95.3&98.6&88.6&~&76.5&74.1&75.2&71.4 \\        
\hline
\end{tabular}
\end{center}
\end{table}
\setlength{\tabcolsep}{1.4pt}
\textbf{Choice of backbone architecture.} As ISP is a backbone-agnostic approach, we show the effectiveness of ISP with different backbones including ResNet \cite{ResNet}, SeResNet  \cite{SENet} and HRNet \cite{HRNet1}. A bilinear upsample layer is added to scale up the final feature maps of ResNet50 and SeResNet50 to the same size of HRNet-W32. As Table \ref{table:backbone} shows, HRNet-W32 obtains the highest performance. We think it is because HRNet maintains high-resolution representations throughout the
network, which could contain more semantic information.
\setlength{\tabcolsep}{4pt}
\begin{table}
\small
\begin{center}
\caption{The ablation studies of different backbone networks on DukeMTMC-reID.}
\label{table:backbone}
\begin{tabular}{ccccc}
\hline\noalign{\smallskip}
backbone & \#params & R-1 & R-5 & mAP \\
\noalign{\smallskip}
\hline
HRNet-W32 & 28.5M & 89.6 & 95.5 & 80.0 \\ 
ResNet50 & 25.6M & 88.7 & 94.9 & 78.9  \\ 
SeResNet50 & 28.1M & 88.8 & 95.2 & 79.2 \\ 
\hline
\end{tabular}
\end{center}
\end{table}
\setlength{\tabcolsep}{1.4pt}

\begin{figure}
\centering
\includegraphics[width=0.9\linewidth, height=4cm]{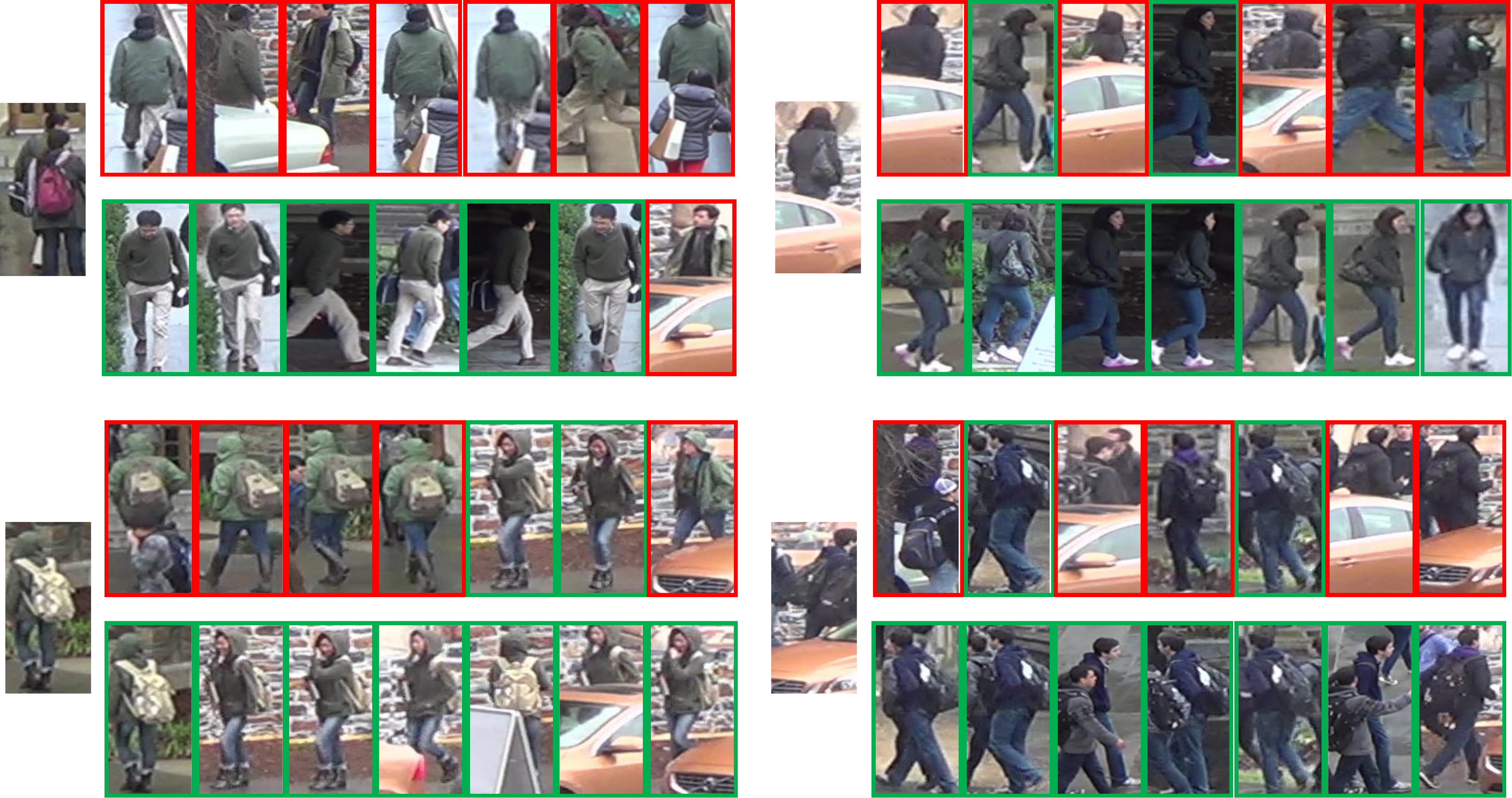}\vspace{-5pt}
\caption{The ranking results of baseline (the first row) and ISP (the second row).}
\label{fig:ranking_result}
\end{figure}

\textbf{The matching results.} We compare the ranking results of baseline and ISP in Figure \ref{fig:ranking_result}, which indicates ISP can well overcome the misalignment problem including part occlusions, inaccurate person detection, and human pose variations. Besides, we can also observe the benefit of identifying the personal belongings.
\section{Conclusion}
In this paper, we propose the identity-guided human semantic parsing method for aligned person re-identification, which can locate both human body parts and personal belongings at pixel-level only with image-level supervision of person identities. 
Extensive experiments validate the superiority of our method.

\noindent\textbf{Acknowledgement}. This work was supported by Key-Area Research and Development Program of Guangdong Province (No.2020B010165001), National Natural Science Foundation of China (No.61772527, 61976210, 61702510), China Postdoctoral Science Foundation No.2019M660859, Open Project of Key Laboratory of Ministry of Public Security for Road Traffic Safety (No.2020ZDSYSKFKT04).


\clearpage
%
%
\bibliographystyle{splncs04}
\bibliography{egbib}
\end{document}